\documentclass[10pt,twocolumn,letterpaper]{article}

\usepackage{iccv}
\usepackage{times}
\usepackage{epsfig}
\usepackage{graphicx}
\usepackage{amsmath}
\usepackage{amssymb}

\usepackage{diagbox}
\usepackage{colortbl}
\usepackage{multirow}
\usepackage[table,xcdraw]{xcolor}
\usepackage[normalem]{ulem}
\useunder{\uline}{\ul}{}
\usepackage{algorithm}
\usepackage{algorithmic}
\usepackage{makecell}
\usepackage{bbding}
\usepackage{subcaption}

\usepackage[breaklinks=true,bookmarks=false]{hyperref}

\iccvfinalcopy % *** Uncomment this line for the final submission

\def\iccvPaperID{****} % *** Enter the ICCV Paper ID here

\ificcvfinal\fi

\begin{document}

%%%%%%%%% TITLE
\title{PanFlowNet: A Flow-Based Deep Network for Pan-sharpening}

\author{Gang Yang$^{1}$, Xiangyong Cao$^{2,3,4}$$\thanks{ corresponding author}$, Wenzhe Xiao$^{2}$,
Man Zhou$^{5}$, Aiping Liu$^{1}$, Xun chen$^{1}$, Deyu Meng$^{2,4}$\\
$^{1}$ University of Science and Technology of China, China\\
$^{2}$ Xi’an Jiaotong University; $^{3}$ School of Computer Science and Technology, Xi’an Jiaotong University\\
$^{4}$ Ministry of Education Key Lab For Intelligent Networks and Network Security, Xi’an Jiaotong University\\
$^{5}$ Nanyang Technological University \\
% {\tt\small \{yg1997, manman\}@mail.ustc.edu.cn, \{caoxiangyong, dymeng\}@mail.xjtu.edu.cn ,wenzhexiao@stu.xjtu.edu.cn,\{aipingl, xunchen\}@ustc.edu.cn}
% \and
% Xiangyong Cao\\
}

\maketitle
\ificcvfinal\thispagestyle{empty}\fi

%%%%%%%%% ABSTRACT
\begin{abstract}
   Pan-sharpening aims to generate a high-resolution multispectral (HRMS) image by integrating the spectral information of a low-resolution multispectral (LRMS) image with the texture details of a high-resolution panchromatic (PAN) image. It essentially inherits the ill-posed nature of the super-resolution (SR) task that diverse HRMS images can degrade into an LRMS image. However, existing deep learning-based methods recover only one HRMS image from the LRMS image and PAN image using a deterministic mapping, thus ignoring the diversity of the HRMS image. In this paper, to alleviate this ill-posed issue, we propose a flow-based pan-sharpening network (\textbf{PanFlowNet}) to directly learn the \textbf{conditional distribution} of HRMS image given LRMS image and PAN image instead of learning a deterministic mapping. Specifically, we first transform this unknown conditional distribution into a given Gaussian distribution by an invertible network, and the conditional distribution can thus be explicitly defined. Then, we design an invertible Conditional Affine Coupling Block (CACB) and further build the architecture of PanFlowNet by stacking a series of CACBs. Finally, the PanFlowNet is trained by maximizing the log-likelihood of the conditional distribution given a training set and can then be used to predict diverse HRMS images. The experimental results verify that the proposed PanFlowNet can generate various HRMS images given an LRMS image and a PAN image. Additionally, the experimental results on different kinds of satellite datasets also demonstrate the superiority of our PanFlowNet compared with other state-of-the-art methods both visually and quantitatively.
\end{abstract}
%%%%%%%%% BODY TEXT
% \section{Introduction}

\begin{figure}[t]
    \centering
    \begin{subfigure}{\linewidth}
    \centering
    % \fbox{\rule{0pt}{2in} \rule{.9\linewidth}{0pt}}
    \includegraphics[width=\linewidth]{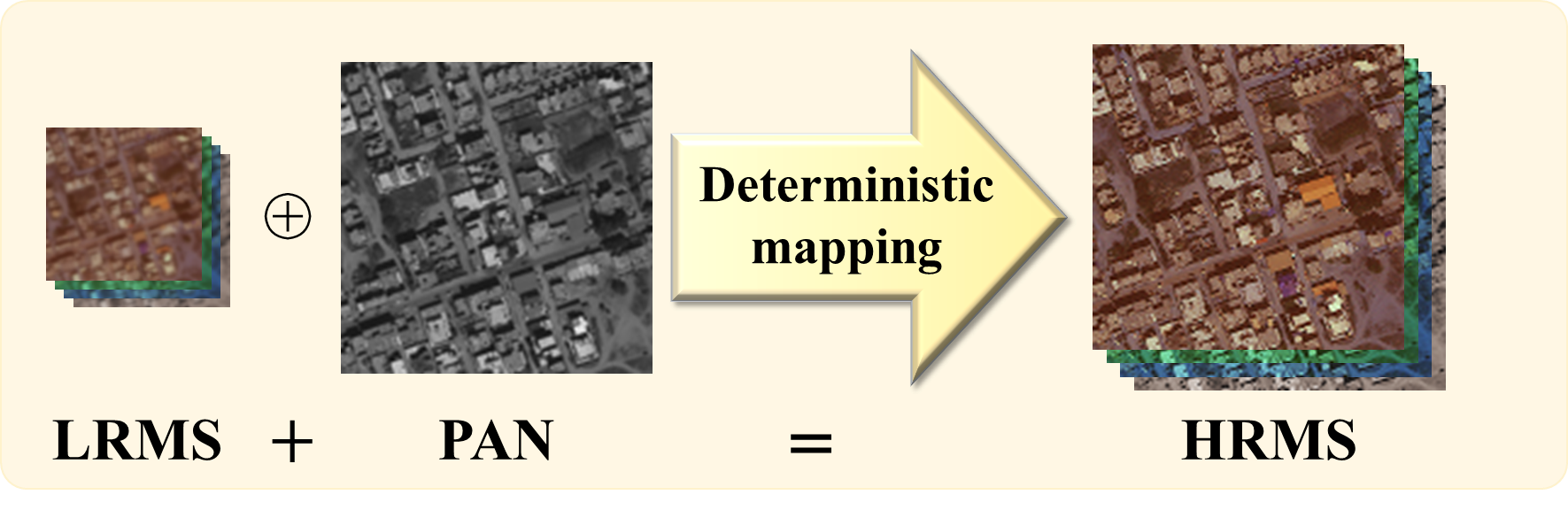}
    \caption{Traditional deep learning-based approaches generate an HRMS image from LRMS and PAN images through deterministic mapping.}
    \label{fig:teaser-a}
  \end{subfigure}
  \vfill
  \begin{subfigure}{\linewidth}
    % \fbox{\rule{0pt}{2in} \rule{.9\linewidth}{0pt}}
    \centering
    \includegraphics[width=\linewidth]{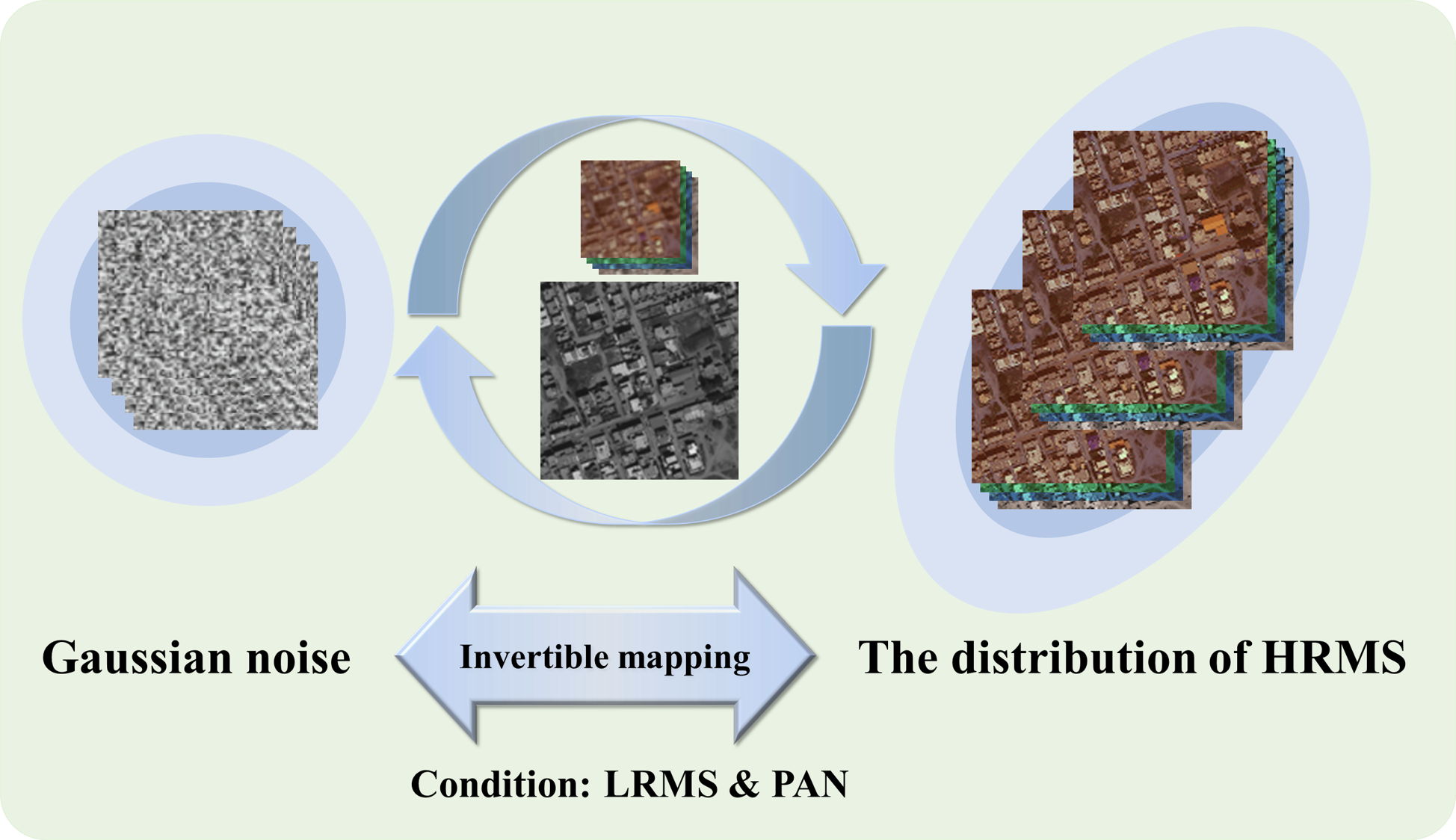}
    \caption{Our proposed PanFlowNet can learn the conditional distribution of HRMS images, and thus it can generate diverse HRMSs from LRMS and PAN images as well as noise.}
    \label{fig:teaser-b}
  \end{subfigure}
  \vspace{-0.8em}
    \caption{Comparison between traditional deep learning-based methods and our proposed PanFlowNet.}
    \vspace{-0.8em}
    \label{fig:teaser}
\end{figure}

%%%%%%%%% BODY TEXT
\vspace{-1.2em}
\section{Introduction}
\label{sec:intro}

With the rapid development of satellite sensors, remote sensing images have become widely used in various applications, such as environmental monitoring~\cite{blaschke2000object}, classification~\cite{cao2018hyperspectral}, and target detection~\cite{ferraris2017robust, zhangtnnls2022}. 
Satellites capture multispectral (MS) and panchromatic (PAN) images simultaneously with complementary information for each modality that PAN images have a high spatial solution~\cite{ghassemian2016review} and MS images contain rich spectral information~\cite{thomas2008synthesis}.
MS sensors reduce the spatial resolution while ensuring spectral richness for MS images~\cite{zhang2021gtp}.
To obtain an MS image with both high spectral and spatial resolution, the pan-sharpening technique that aims to fuse the MS and PAN images has attracted a large amount of attention.

The past decades have witnessed the explosive growth of research works in the pan-sharpening field. In terms of the quality of the generated fusion results,  the focuses have been mainly on model-based~\cite{thomas2008synthesis, vivone2014critical, meng2019review} and deep learning (DL)-based~\cite{deng2020detail, fu2020deep, wald1997fusion, wei2017boosting, yang2017pannet, chenijcai2022} methods. Model-based methods usually optimize a mathematical model that preserves spectral and spatial information, and most of them follow the assumption that the PAN image (or its gradient) can be modelled as a linear combination among all bands (or their gradients) of high-resolution multispectral (HRMS) images. However, they are highly dependent on the assumptions about the relationship between HRMS and PAN images~\cite{P+XS2006}. Unfortunately, previous work did not accurately establish this relationship, which limits the further improvement of pan-sharpening. Besides, the model-based methods are challenging in optimization, limiting their practical applications. 

In the era of deep learning, convolutional neural networks (CNN) have emerged as a significant tool for pan-sharpening. CNN-based methods train the network by minimizing the distance between the fused result and the HRMS reference image. Because of the strong nonlinear fitting ability of neural networks, this kind of method always achieves excellent performance. However, pan-sharpening is essentially an ill-posed problem since a given LRMS image can be degraded from infinitely many compatible HRMS images. This poses severe challenges when designing DL-based pan-sharpening approaches. Although existing CNN-based methods can obtain excellent results, they only learn a deterministic mapping from LRMS and PAN images to HRMS images, as shown in Fig.~\ref{fig:teaser-a}, and thus the ill-posed issue is not well addressed. 
%See Fig.~\ref{fig:teaser-a} for an illustration. When trained with many-to-one mapping between the high resolution images and the low resolution images, existing methods tend to adopt the average of all possible HR images, thus suffering from unreal SR images. These methods only predict a single SR output through a deterministic mapping, which does not fully account for the ill-posed nature of the SR problem.

To solve the above issues and generate more diverse realistic images, in this paper, we propose a novel neural architecture for pan-sharpening, called \textbf{PanFlowNet}, which directly learns the conditional distribution of the HRMS image given the input LRMS image and PAN image instead of learning a deterministic mapping. Specifically, we first transform a sample of the unknown conditional distribution into a sample of a given Gaussian distribution using an invertible network. Thus the \textbf{\textit{conditional distribution}} can be explicitly defined by the product of the Gaussian distribution and the determinant of the Jacobian matrix. To build the invertible network, we first design an invertible Conditional Affine Coupling Block (CACB) and then stack a series of CACBs to construct the network architecture of PanFlowNet. Finally, our PanFlowNet can be trained by minimizing the negative log-likelihood of the conditional distribution on a training set. Once the training is finished, we can generate diverse HRMSs by inputting the LRMS and PAN images as well as different noise samples of the given Gaussian distribution into the PanFlowNet, as shown in Fig.~\ref{fig:teaser-b}.

%Fig.~\ref{fig:teaser-b} for an illustration. We address the task of pan-sharpening guided by LR MS and PAN when the generative model generates the distribution of samples by passing random noise through a flow-based deep network.
%In order to maintain the reversibility of the flow model and to take full advantage of the relationship between the channels of the features, we introduce Conditional Affine Coupling Block as the basic structure in the conditional generation model.
% \subsection{Motivation}
% We formulate the pan-sharpening as a generative problem with the distribution of HR MS samples generated by random noise guided by LRMS and PAN.
% This approach explicitly addresses the ill-posed nature of the SR problem by aiming to capture the full diversity of possible SR images from the natural image manifold.
% To this end, we design a conditional normalizing flow architecture, allowing us to learn rich distributions using exact log-likelihood based training.
In summary, the contributions of our work are as follows:
\begin{itemize}
    \item We propose a flow-based deep network (i.e., PanFlowNet) for pan-sharpening. This network can accurately learn the conditional distribution of HRMS images given the corresponding LRMS and PAN images. To the best of our knowledge, this is the first attempt to learn an explicit distribution by employing the generative flow model for the pan-sharpening task.
    \item The proposed PanFlowNet can generate diverse HRMSs given the LRMS and PAN images as well as the Gaussian noise sample and thus can alleviate the ill-posed issue to some extent. Besides, the generated HRMS images are diverse since each HRMS image focuses on a different detailed part of the ground truth. 
    \item We extend the vanilla flow model to a probabilistic multi-conditional flow model to adapt to the multi-conditionality of the pan-sharpening task. Extensive experiments over different satellite datasets demonstrate that our method can outperform existing state-of-the-art approaches both visually and quantitatively.
\end{itemize}

%\begin{figure}[t]
%  \centering
%  \fbox{\rule{0pt}{2.4in} \rule{0.9\linewidth}{0pt}}
   % \includegraphics[width=\linewidth]{./figs/teaser.png}

%   \caption{}
%   \label{fig:teaser-c}
%\end{figure}

\section{Related work}
\label{sec:related}

\subsection{Classic pan-sharpening methods}

The traditional methods of pan-sharpening can be classified into three main categories: component substitution (CS)-~\cite{carper1990use, shah2008efficient, gillespie1987color}, multi-resolution analysis (MRA)-~\cite{ATWT1999, vivone2014critical, HPF, DWT1989}, and variational optimization (VO)-~\cite{LGC2019, ADMM2021, deng2018variational, P+XS2006, SIRF2015} based methods. The common CS methods~\cite{carper1990use, shah2008efficient, gillespie1987color} project the original MS image into a transform domain and then replace the separated spatial components with PAN images. The typical MRA methods~\cite{HPF, DWT1989} inject the spatial details extracted by the multiresolution decomposition techniques from PAN images into the up-sampled MS images. The VO methods~\cite{P+XS2006, SIRF2015} are concerned because of the fine fusion effects on pan-sharpening. In addition, some hybrid methods take advantage of multiple methods to complement each other~\cite{zhang2019pan, kwan2018super}.
Most model-based methods assume that the PAN image (or its gradient) can be modelled as a linear combination among all bands (or their gradients) of the HRMS image. 
However, this reduces the intensity fidelity of the HRMS image since various sensors mounted on satellites have extremely diverse response characteristics to objects~\cite{zhang2021gtp}.
% But because the response characteristics of different sensors mounted on satellites to objects are very different, this reduces the intensity fidelity of the HRMS image~\cite{zhang2021gtp}.

\subsection{Deep learning based pan-sharpening methods}

Due to the highly nonlinear fitting capacity of the convolutional neural network, PNN~\cite{masi2016} models the relationship between PAN, LRMS, and HRMS images using three convolutional layers, achieving a significant improvement compared with other classical methods. Inspired by PNN, a large number of CNN-based pan-sharpening studies~\cite{caoUnfolding, ma2021pan, xu2020sdpnet} have emerged recently. 
% However, most of the early works stack the existing CNN modules borrowed from other visual tasks. 
For instance, PANNet~\cite{yang2017pannet} utilizes ResNet's residual learning module, MSDCNN~\cite{yuan2018multiscale} adds multi-scale modules based on residual connection, SRPPNN~\cite{cai2020super} refers to the design idea of SRCNN~\cite{dong2015image}, and Wang~\etal~\cite{wang2021ssconv} adopted U-shaped network. 
Moreover, WSDFNet~\cite{jin2021weighted} propagates shallow features scaled by adaptive skip weightier, and Ma~\etal~\cite{ma2020pan} proposes an unsupervised framework based on GAN. 
% Observing that the same object in MS and PAN is not always aligned, Li~\emph{et al.}~\cite{SIPSA-Net2021} design a SIPSA-Net with a feature alignment module which can align features from PAN and LR MS images. Wu~\emph{et al.}~\cite{Wu_2021_ICCV} utilize multiple parallel branches to integrate features of different scales into the backbone network to improve performance.
Additionally, some model-driven CNN models with clear physical meaning emerge, such as MHNet~\cite{MHNet2019}, Proximal PanNet~\cite{cao2022proximal}, PanCSC-Net~\cite{caoUnfolding}, MADUN~\cite{zhou2022memory}, GPPNN~\cite{Xu_2021_CVPR}. Although all these DL-based pansharpening approaches achieve excellent performance, they all only learn a deterministic mapping from the LRMS and PAN image to the HRMS image, thus ignoring the ill-posed issue of the pansharpening task. 

\subsection{Flow-based methods}

Flow-based generative models have shown an excellent ability to explicitly learn the probability density function of data.
% via a sequence of invertible transformations.
A sequence of invertible transformations generally constructs them to map a base distribution to a complex one~\cite{papamakarios2021normalizing, ho2019flow++, rezende2015variational, kobyzev2020normalizing}. 
Several unconditional generative flow models have emerged that extend the early flow models to multiscale architectures with split couplings that allow for efficient inference and sampling. For example, 
Dinh~\etal~\cite{dinh2014nice} proposes to stack non-linear additive coupling and other transformation layers as the flow model NICE. Inspired by NICE, Dinh~\etal~\cite{dinh2016density} propose RealNVP, which upgrades additive coupling to affine coupling without loss of invertibility and achieves better performance. After that, Kingma~\etal~\cite{kingma2018glow} propose 1×1 convolution to replace the fixed permutation layer in RealNVP and succeed in synthesizing realistic-looking images. Likewise, various conditional flow models have appeared aiming at conditional image synthesis~\cite{sun2019dual, ardizzone2019guided}. Lugmayr~\etal proposed SRFlow~\cite{lugmayr2020srflow} to generate diverse high-resolution images conditioned on low-resolution ones. Abdal~\etal~\cite{abdal2021styleflow} sampled latent vectors based on given attributes and fed the vectors to the StyleGANgenerator to synthesize high-quality images. Compared with the aforementioned methods, especially SRFlow~\cite{lugmayr2020srflow}, our approach has several differences. Firstly, our approach is specifically proposed for the pan-sharpening task. Secondly, unlike the flow models for traditional image inverse problems that have no extra guided image information, e.g., SRFlow, the flow model of pansharpening requires embedding the guided PAN image information and thus poses an issue of how to inject the detailed texture information of PAN image into the network. %architecture. 

\section{Proposed method}\label{sec:methods}
In this section, we first introduce our proposed probabilistic flow model for pan-sharpening in detail. Then, we design a network to implement this model.

\begin{figure*}
  \centering
  \vspace{-0.5em}
  \includegraphics[width=0.88\linewidth]{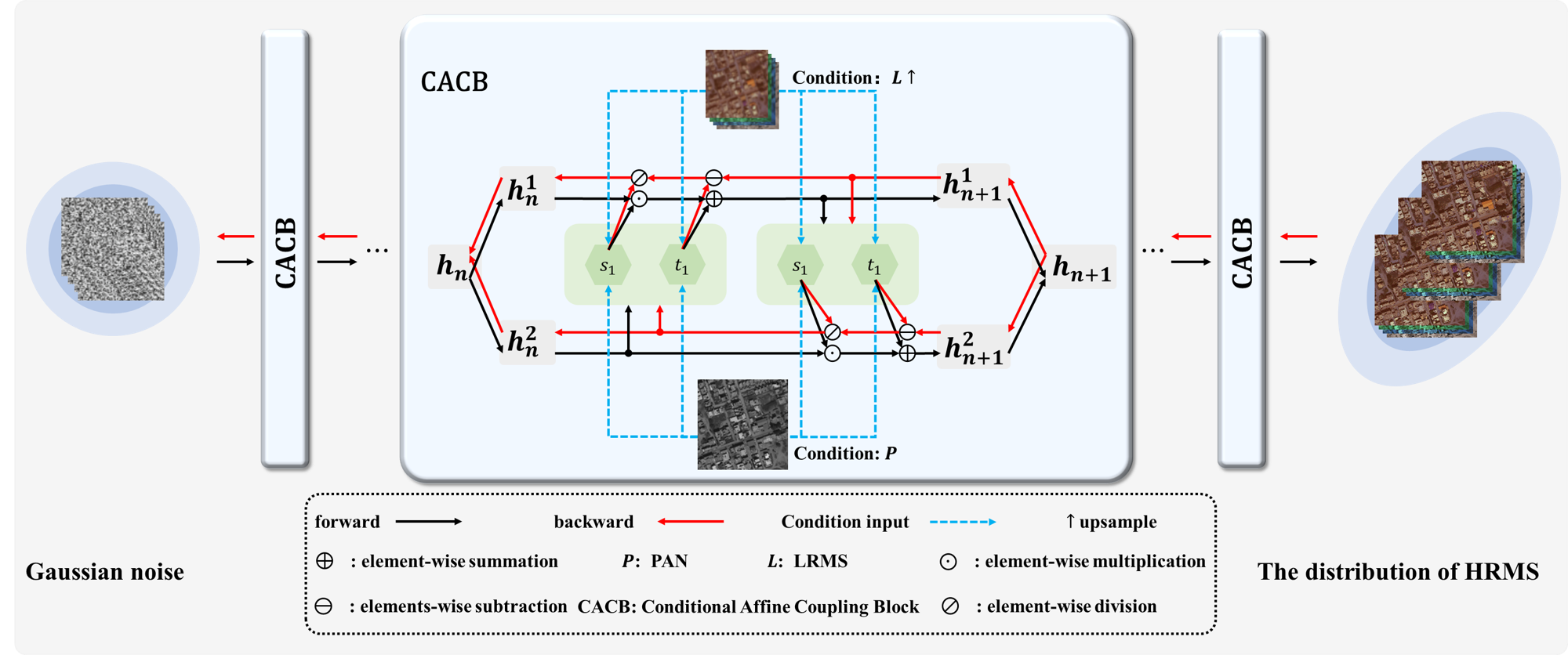}
%   \begin{subfigure}{0.48\linewidth}
%     % \fbox{\rule{0pt}{2in} \rule{.9\linewidth}{0pt}}
%     \includegraphics[width=\linewidth]{cvpr2023-author_kit-v1_1-1/latex/figs/arch.png}
%     \caption{An example of a subfigure.}
%     \label{fig:arch-a}
%   \end{subfigure}
%   \hfill
%   \begin{subfigure}{0.48\linewidth}
%     % \fbox{\rule{0pt}{2in} \rule{.9\linewidth}{0pt}}
%     \includegraphics[width=\linewidth]{cvpr2023-author_kit-v1_1-1/latex/figs/cacb.png}
%     \caption{Another example of a subfigure.}
%     \label{fig:arch-b}
%   \end{subfigure}
  \caption{The network architecture of our PanFlowNet consists of a series of invertible Conditional Affine Coupling Blocks (CACBs). The PanFlowNet can directly learn the distribution of HRMS images from Gaussian noise conditioned on LRMS and PAN images.}
  \vspace{-1.2em}
  \label{fig:arch}
\end{figure*}

\subsection{Probabilistic flow model for pan-sharpening}

The goal of pan-sharpening is to recover the high-resolution multispectral (HRMS) image $\mathbf{H}\in\mathbb{R}^{H\times W\times B}$ from a given low-resolution multispectral (LRMS) image $\mathbf{L}\in\mathbb{R}^{h\times w\times B}$ under the guidance of a high resolution panchromatic (PAN) image $\mathbf{P}$, where $h=H/s, w=W/s$, and $s$ is a resolution factor. 

As aforementioned, the pan-sharpening task is essentially an ill-posed problem since the LRMS image may be degraded from an infinite amount of HRMSs. However, most current deep learning-based pan-sharpening approaches learn a deterministic mapping $f: (\mathbf{L},\mathbf{P})\longmapsto\mathbf{H}$, which receives the LRMS image $\mathbf{L}$ and the PAN image $\mathbf{P}$ as input and outputs only one possible HRMS image $\mathbf{H}$. To alleviate the ill-posed issue, this work aims to learn the conditional distribution of the HRMS image $\mathbf{H}$, i.e., $P_{\mathbf{H}|\mathbf{L},\mathbf{P}}(\mathbf{H}|\mathbf{L},\mathbf{P};\boldsymbol{\theta})$, given the LRMS image $\mathbf{L}$ and the PAN image $\mathbf{P}$, which is a more difficult task since the conditional distribution model can generate infinite possible HRMS images, instead of just predicting a single HRMS image output by a deterministic mapping. Next, we will propose a probabilistic flow method to learn the conditional distribution $P_{\mathbf{H}|\mathbf{L},\mathbf{P}}(\mathbf{H}|\mathbf{L},\mathbf{P};\boldsymbol{\theta})$ given an LRMS-PAN-HRMS training set $\mathcal{D}=\{(\mathbf{L}_{j},\mathbf{P}_{j},\mathbf{H}_{j})\}_{j=1}^{m}$. 

Since the conditional distribution $P_{\mathbf{H}|\mathbf{L},\mathbf{P}}(\mathbf{H}|\mathbf{L},\mathbf{P};\boldsymbol{\theta})$ is unknown, we thus resort to the probabilistic flow model, which uses an invertible function $f_{\boldsymbol{\theta}}(\cdot)$ to parametrize the conditional distribution. In this conditional setting, $f_{\boldsymbol{\theta}}(\cdot)$ can map a LRMS-PAN-HRMS image pair to a latent variable $\mathbf{z}$, namely
\begin{eqnarray}\label{forward}
\mathbf{z}=f_{\boldsymbol{\theta}}(\mathbf{H};\mathbf{L},\mathbf{P}),
\end{eqnarray}
where $f_{\boldsymbol{\theta}}(\mathbf{H};\mathbf{L},\mathbf{P})$ is required to be invertible for the first argument $\mathbf{H}$ given the LRMS image $\mathbf{L}$ and the PAN image $\mathbf{P}$. Therefore, the HRMS image $\mathbf{H}$ can then be exactly obtained from the latent variable $\mathbf{z}$ as
\begin{eqnarray}\label{sample}
\mathbf{H}=f_{\boldsymbol{\theta}}^{-1}(\mathbf{z};\mathbf{L},\mathbf{P}),~~\mathbf{z}\sim P_{\mathbf{z}}(\cdot),
\end{eqnarray} 
where $\mathbf{z}$ is a sample of the latent variable distribution $P_{\mathbf{z}}(\cdot)$. For simplicity, the latent variable $\mathbf{z}$ is always assumed to be a simple Gaussian distribution $\mathbf{z}\sim P_{\mathbf{z}}(\mathbf{z})=\mathcal{N}(\mathbf{z}|\mathbf{0},\mathbf{I}_{s})$, where $\mathbf{I}_{s}$ is the identity matrix, $s$ is the dimension of $\mathbf{z}$, and $s$ equals to $HWB$ in order to guarantee $f_{\boldsymbol{\theta}}$ to be invertible. In this setting, the probability density function $P_{\mathbf{H}|\mathbf{L},\mathbf{P}}(\mathbf{H}|\mathbf{L},\mathbf{P};\boldsymbol{\theta})$ can then be accurately defined by using the change-of-variables formula, namely 
\begin{equation}\label{prob}
	P_{\mathbf{H}|\mathbf{L},\mathbf{P}}(\!\mathbf{H}|\!\mathbf{L},\!\mathbf{P};\!\boldsymbol{\theta})\!=\!P_{\mathbf{z}}(f_{\boldsymbol{\theta}}(\!\mathbf{H};\!\mathbf{L},\!\mathbf{P}\!))\left|\!\det\!\frac{\partial f_{\boldsymbol{\theta}}(\mathbf{H};\mathbf{L},\mathbf{P})}{\partial\mathbf{H}}\right|,  
\end{equation}
where $\frac{\partial f_{\boldsymbol{\theta}}(\mathbf{H};\mathbf{L},\mathbf{P})}{\partial\mathbf{H}}$ is the Jacobian matrix of function $f_{\boldsymbol{\theta}}(\cdot)$ at $\mathbf{H}$ and ${\rm{det}(\cdot)}$ is the determinant function. In this work, 
we choose $f_{\boldsymbol{\theta}}(\cdot)$ such that its determinant of the Jacobian is easily computed. Specifically, we utilize an inverse neural network (INN) to implement $f_{\boldsymbol{\theta}}(\cdot)$, and the detailed design of the INN is presented in the next section. Based on Eq.~(\ref{prob}), we can learn the parameter of $f_{\boldsymbol{\theta}}$ by minimizing the negative log-likelihood (NLL) of training pair $(\mathbf{L},\mathbf{P},\mathbf{H})$ as follows:
\begin{eqnarray}\label{nll}
	\begin{split}
		&\mathcal{L}(\boldsymbol{\theta};\mathbf{L},\mathbf{P},\mathbf{H})=-\log P_{\mathbf{H}|\mathbf{L},\mathbf{P}}(\mathbf{H}|\mathbf{L},\mathbf{P};\boldsymbol{\theta}) \\
		&= -\log P_{\mathbf{z}}(f_{\boldsymbol{\theta}}(\mathbf{H};\mathbf{L},\mathbf{P}))-\log\left|\det\frac{\partial f_{\boldsymbol{\theta}}(\mathbf{H};\mathbf{L},\mathbf{P})}{\partial\mathbf{H}}\right|.
	\end{split}
\end{eqnarray}

Further, to guarantee the second term in Eq.~(\ref{nll}) tractable, we decompose $f_{\boldsymbol{\theta}}$ into a sequence of $N$ invertible layers, i.e., $f_{\boldsymbol{\theta}}=f_{\boldsymbol{\theta}}^{N}f_{\boldsymbol{\theta}}^{N-1}\cdots f_{\boldsymbol{\theta}}^{1}$, where $f_{\boldsymbol{\theta}}^{n}$ is the $n_{th}$ invertible layer, which receives feature $\mathbf{h}_{n}$ of the preivous layer as input and generates $\mathbf{h}_{n+1}$ as output, i.e., $\mathbf{h}^{n+1}=f_{\boldsymbol{\theta}}^{n}(\mathbf{h}^{n};\mathbf{L},\mathbf{P})$. Based on Eq.~(\ref{forward}), we can easily know that $\mathbf{h}^{1}=\mathbf{H}$ and $\mathbf{h}^{N+1}=\mathbf{z}$. Additionally, we encode the information of LRMS image $\mathbf{L}$ and PAN image $\mathbf{P}$ into each invertible layer $f_{\boldsymbol{\theta}}^{n}$ by regarding them as conditional input, which can compensate for detail and structure information to each layer. 

By utilizing the chain rule and the multiplicative property of the determinant, we can compute the NLL objective in Eq.~(\ref{nll}) as follows:
\begin{eqnarray}\label{nll_ml}
	\begin{split}
		&\mathcal{L}(\boldsymbol{\theta};\mathbf{L},\mathbf{P},\!\mathbf{H}) \\
		&\!=\! -\log P_{\mathbf{z}}\!(f_{\boldsymbol{\theta}}(\!\mathbf{H};\!\mathbf{L},\!\!\mathbf{P}))-\sum_{n=0}^{N\!-\!1}\!\!\log\left|\!\det\!\frac{\partial f_{\boldsymbol{\theta}}^{n}(\mathbf{h}^{n};\mathbf{L},\mathbf{P})}{\partial\mathbf{h}^{n}}\right|.
	\end{split}
\end{eqnarray}
Therefore, the final objective function on the training set $\mathcal{D}$ is defined as $\mathcal{L}_{final}(\boldsymbol{\theta})=\sum_{j=1}^{m}\mathcal{L}(\boldsymbol{\theta};\mathbf{L}_{j},\mathbf{P}_{j},\mathbf{H}_{j})$. To ensure each layer invertible and fast computation of the log-determinant of the Jacobian $\frac{\partial f_{\boldsymbol{\theta}}^{n}}{\partial\mathbf{h}^{n}}$, we need to carefully design the network architecture of each layer. This will be discussed in the next section. 

% Once the optimal parameter $\boldsymbol{\theta}_{*}$ is learned, we can directly sample the HRMS image from $P_{\mathbf{H}|\mathbf{L},\mathbf{P}}(\mathbf{H}|\mathbf{L},\mathbf{P};\boldsymbol{\theta}_{*})$ based on Eq.(\ref{sample}) by applying the inverse network $f_{\boldsymbol{\theta}}^{-1}$ to a sample $\mathbf{z}$, i.e.,  $\mathbf{H}=f_{\boldsymbol{\theta}_{*}}^{-1}(\mathbf{z};\mathbf{L},\mathbf{P}), \mathbf{z}\sim \mathcal{N}(\mathbf{z}|\mathbf{0},\mathbf{I}_{s})$. 

Once the optimal parameter $\theta^{*}$ of the invertible network $f_{\theta}$ is learned, we can sample an HRMS from $P_{\mathbf{H}|\mathbf{L},\mathbf{P}}(\mathbf{H}|\mathbf{L},\mathbf{P};\boldsymbol{\theta}_{*})$ as follows: 
\begin{eqnarray}
    \mathbf{H}=f_{\boldsymbol{\theta}_{*}}^{-1}(\mathbf{z};\mathbf{L},\mathbf{P}),~~\mathbf{z}\sim \mathcal{N}(\mathbf{z}|\mathbf{0},\mathbf{I}_{s}). 
\end{eqnarray}
Since we can sample infinite HRMS images, a vital issue is posed that which sample we should choose in practical application. In this paper, we propose a \textbf{\textit{maximum probability criterion}}, i.e., the HRMS image corresponding to the maximum probability of $P_{\mathbf{H}|\mathbf{L},\mathbf{P}}(\mathbf{H}|\mathbf{L},\mathbf{P};\boldsymbol{\theta}_{*})$ calculated by Eq.(\ref{sample}) is selected. The proposed criterion is used in all the experiments.

\subsection{Network architecture}\label{sec:architecture}

\subsubsection{Overall network architecture} \label{sec:overall_network}
The model must span a variety of possible HRMS images instead of just predicting a single HRMS output. Our intention is to learn the parameters $\boldsymbol \theta$ of the distribution in a data-driven manner, given a training set.

In this section, we follow the design of the Probabilistic Flow Model to produce an invertible neural network (INN) with Conditional Affine Coupling Blocks (CACBs). Specifically, we build \textbf{PanFlowNet} by stacking a series of invertible layers. As shown in Fig.~\ref{fig:arch}, PanFlowNet consists of several flow blocks, and each flow block is composed of a reversible CACB. A CACB corresponds to one step of the transformation process $f_{\boldsymbol{\theta}}^{n}$. For each flow layer, the transformation process is considered as follows.

Let $\mathbf{h}_{n}$ be a latent feature variable in a series of invertible transformations. The goal of flow layer $f_{\boldsymbol{\theta}}^{n}$ is to generate $\mathbf{h}_{n+1}$ with the guidance of LRMS and PAN images as
\begin{align}
    \mathbf{h}_{n+1}\;=\;f_{\boldsymbol{\theta}}^{n}(\mathbf{h}_{n}; \mathbf{L}, \mathbf{P}).  \label{eq:h_n}
\end{align}
The inverse process of flow layer ${g_{\boldsymbol{\theta}}^{n}}$ takes $\mathbf{h}_{n+1}$ as input and uses LRMS and PAN as conditions to generate $\mathbf{h}_{n}$, and the process can be expressed as follows:
\begin{align}
    \mathbf{h}_{n}\;=\;g_{\boldsymbol{\theta}}^{n}(\mathbf{h}_{n+1}; \mathbf{L}, \mathbf{P}). \label{eq:g_n}
\end{align}
We follow the design of Eq.~\ref{eq:h_n} and Eq.~\ref{eq:g_n} to provide a CACB as $f_{\boldsymbol{\theta}}^{n}$ and the reverse as ${g_{\boldsymbol{\theta}}^{n}}$. In the next subsection, we will present the structural design of the CACB.

\subsubsection{Conditional affine coupling block} \label{sec:coupling-block}

\begin{figure}
  \centering
  % \fbox{\rule{0pt}{2.4in} \rule{0.9\linewidth}{0pt}}
  \includegraphics[width=\linewidth]{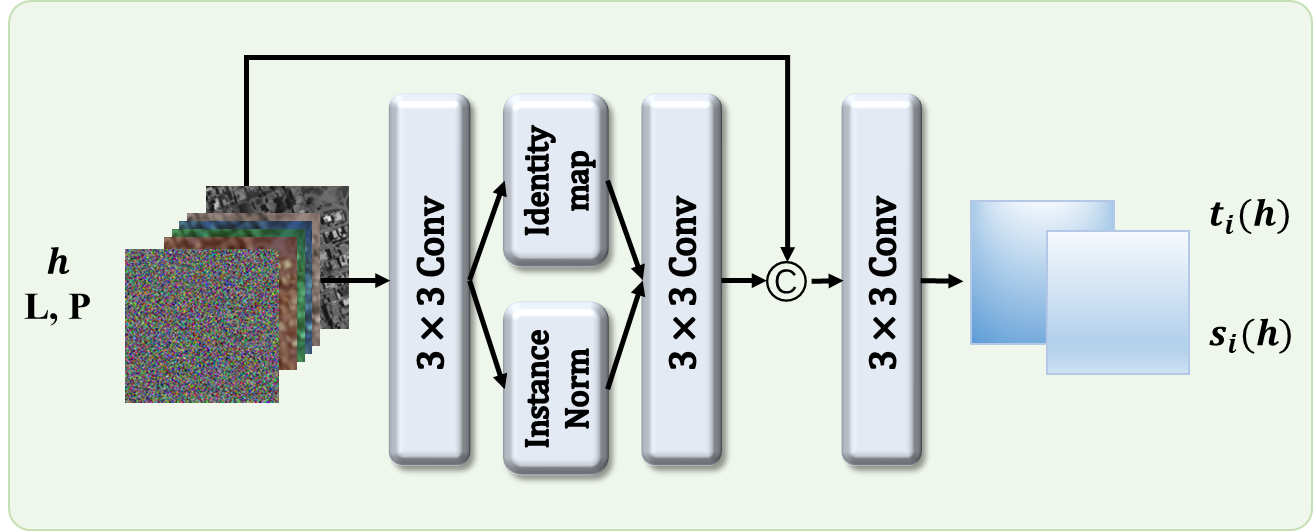}
  \caption{HIN Block is used to implement $s_i(\cdot)$ and $t_i(\cdot)$}
  \vspace{-0.8em}
  \label{fig:hin}
\end{figure}

The network architecture of the invertible layer requires careful design in order to ensure well-conditioned reversibility and a tractable Jacobian determinant. This challenge was first addressed in~\cite{dinh2014nice, dinh2016density} and has recently inspired significant interest. Our method is an extension of the affine coupling block architecture established in~\cite{dinh2016density}. As shown in Fig.~\ref{fig:arch}, each flow block is a reversible block consisting of two complementary affine coupling layers, which splits its input $\mathbf{h}_{n}$ into two parts, i.e., $\mathbf{h}_{n}$=[$\mathbf{h}_{n}^1$, $\mathbf{h}_{n}^2$], and applies affine transformations with coefficients $exp(s_i)$ and $t_i, i=1, 2$ to them. Specifically, the affine transformation is defined as follows:
\begin{align}
    \mathbf{h}_{n+1}^{1} &=\mathbf{h}_{n}^1 \odot \exp \left(s_{1}\left(\mathbf{h}_{n}^2\right)\right)+t_{1}\left(\mathbf{h}_{n}^2\right),  \label{eq:forward_1}\\
   \mathbf{h}_{n+1}^{2} &=\mathbf{h}_{n}^2 \odot \exp \left(s_{2}\left(\mathbf{h}_{n+1}^1\right)\right)+t_{2}\left(\mathbf{h}_{n+1}^1\right), \label{eq:forward_2}
\end{align}
where $\odot$ is element-wise multiplication. This affine transformation has a triangular Jacobian matrix, and thus its determinant is easy to calculate. Additionally, the output $[\mathbf{h}_{n+1}^{1}, \mathbf{h}_{n+1}^{2}]$ are concatenated again and then passed to the next coupling block. The internal functions $s_i(\cdot)$ and $t_i(\cdot)$ can be represented by arbitrary neural networks and are only evaluated in the forward direction. This affine transformation in Eq.~(\ref{eq:forward_1}) and Eq.~(\ref{eq:forward_2}) are easily to  invertible, namely
\begin{align}
    \mathbf{h}_{n}^2\;&=\;\left(\mathbf{h}_{n+1}^{2}-t_{2}\left(\mathbf{h}_{n+1}^{1}\right)\right) \oslash \exp \left(s_{2}\left(\mathbf{h}_{n+1}^{1}\right)\right), \\
    \mathbf{h}_{n}^1\;&=\;\left(\mathbf{h}_{n+1}^{1}-t_{1}\left(\mathbf{h}_{n}^{2}\right)\right) \oslash \exp \left(s_{1}\left(\mathbf{h}_{n}^{2}\right)\right),
\end{align}
where $\oslash$ is element-wise division. As shown in~\cite{dinh2016density}, the logarithm of the Jacobian determinant for such a coupling block is simply the sum of $s_1(\cdot)$ and $s_2(\cdot)$ over image dimensions. In the conditional setting that $\mathbf{L}$ and $\mathbf{P}$ are regarded as conditional input of $s_{i}(\cdot)$ and $t_{i}(\cdot)$, $i=1,2$, the affine transformation is still invertible since its invertibility is not related with the sub-networks $s_j(\cdot)$ and $t_j(\cdot)$. Thus, we generate the input for $s(\cdot)$ and $t(\cdot)$ by concatenating the condition data $\mathbf{L}$ and $\mathbf{P}$ with the latent feature $\mathbf{h}$, which will not affect the invertibility of this affine transformation. Fig.~\ref{fig:hin} shows the conditional affine coupling layer, which is an \textbf{\textit{extension}} of the affine coupling layer presented above. In our implementation, $s_i(\cdot)$ and $t_i(\cdot)$ in the $f_{\boldsymbol{\theta}}^{n}$ are implemented  by HIN Block~\cite{chen2021hinet}.

\begin{figure*}[ht]
  \centering
  \includegraphics[width=0.72\linewidth]{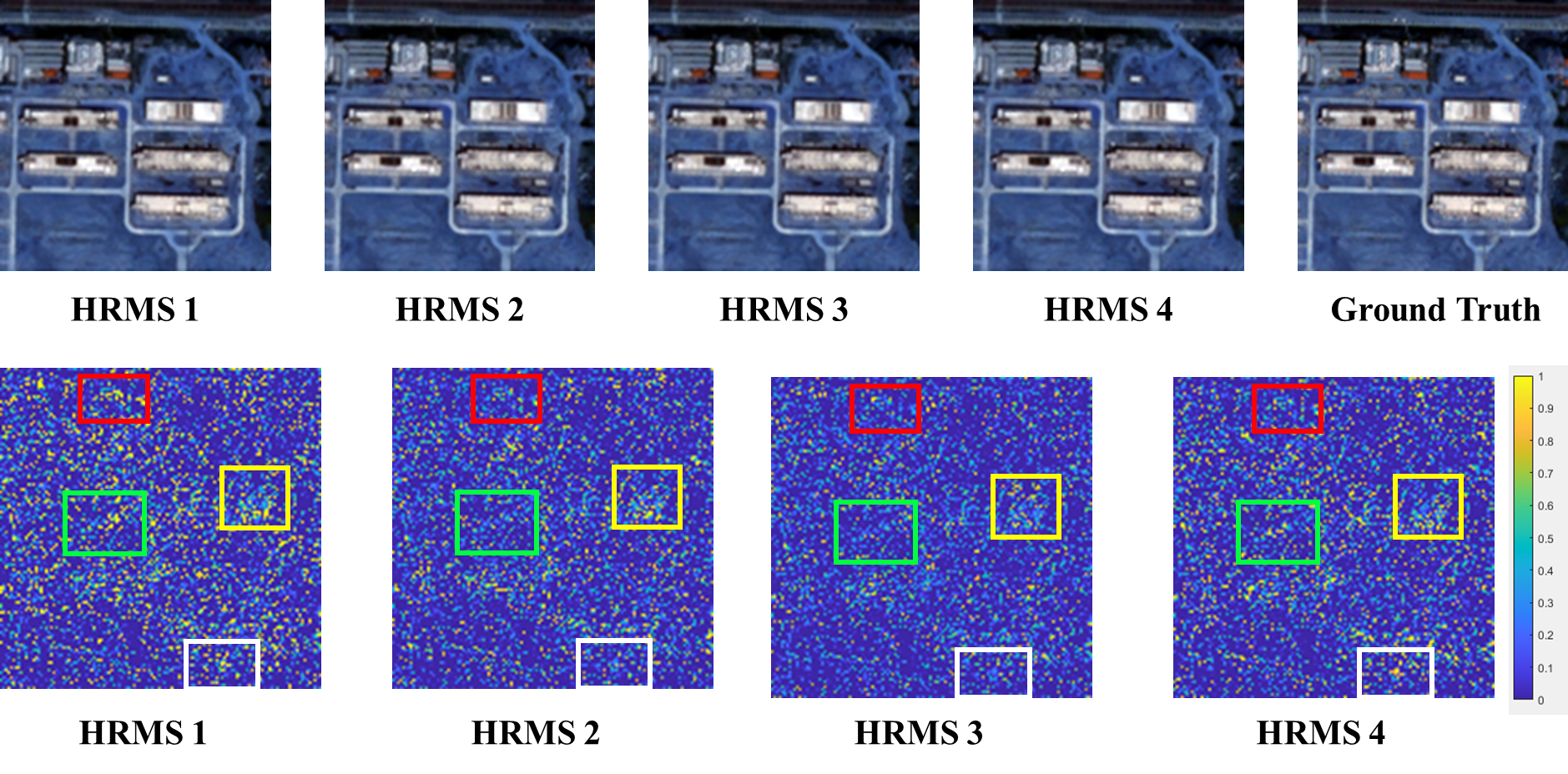}
  \vspace{-0.8em}
  \caption{The visualization results are used to validate the effectiveness of our proposed PanFlowNet. The first row visualizes different HRMS images generated from different noises and gives LRMS and PAN images on the WorldView-II dataset. The second row visually shows the differences in the detailed parts that each HRMS image focuses on ground truth.}
  \label{fig:assess_motivation}
  \vspace{-0.8em}
\end{figure*}

\begin{table*}[ht]
\caption{Experimental results of all the competing methods on the three benchmark datasets. The best and the second best values are highlighted in \textcolor{red}{\textbf{bold}} and \textcolor{blue}{\ul underline}, respectively.}
\vspace{-0.6em}
\label{tab:cmp_sota}
\renewcommand\arraystretch{1.2}
\resizebox{\textwidth}{!}{%
\begin{tabular}{c|c|cccc|cccc|cccc}
\hline
\rowcolor[HTML]{ECF4FF} 
\cellcolor[HTML]{ECF4FF} &
\cellcolor[HTML]{ECF4FF} &
\multicolumn{4}{c|}{\cellcolor[HTML]{ECF4FF}WorldView II} &
\multicolumn{4}{c|}{\cellcolor[HTML]{ECF4FF}WorldView III} &
\multicolumn{4}{c}{\cellcolor[HTML]{ECF4FF}GaoFen2} \\ \cline{3-14} 
\rowcolor[HTML]{ECF4FF} 
\multirow{-2}{*}{\cellcolor[HTML]{ECF4FF}Methods} &
\multirow{-2}{*}{\cellcolor[HTML]{ECF4FF}Params} &
\cellcolor[HTML]{ECF4FF}PSNR $\uparrow$ &
\cellcolor[HTML]{ECF4FF}SSIM $\uparrow$ &
\cellcolor[HTML]{ECF4FF}SAM $\downarrow$ &
\cellcolor[HTML]{ECF4FF}ERGAS $\downarrow$ &
PSNR $\uparrow$ &
SSIM $\uparrow$ &
SAM $\downarrow$ &
ERGAS $\downarrow$ &
PSNR $\uparrow$ &
SSIM $\uparrow$ &
SAM $\downarrow$ &
ERGAS $\downarrow$ \\ \hline
SFIM   & -      & 34.1297 & 0.8975 & 0.0439  & 2.3449 & 21.8212 & 0.5457 & 0.1208 & 8.973  & 36.906  & 0.8882 & 0.0318 & 1.7398 \\
Brovey & -      & 35.8646 & 0.9216 & 0.0403  & 1.8238 & 22.5060  & 0.5466 & 0.1159 & 8.2331 & 37.7974 & 0.9026 & 0.0218 & 1.372  \\
GS     & -      & 35.6376 & 0.9176 & 0.0423  & 1.8774 & 22.5608 & 0.547  & 0.1217 & 8.2433 & 37.226  & 0.9034 & 0.0309 & 1.6736 \\
IHS    & -      & 32.1601 & \textcolor{red}{\textbf{0.9812}} & 10.3010 & 26.40  & 22.5579 & 0.5354 & 0.1266 & 8.3616 & 38.1754 & 0.9100 & 0.0243 & 1.5336 \\
GFPCA  & -      & 34.5581 & 0.9038 & 0.0488  & 2.1411 & 22.3344 & 0.4826 & 0.1294 & 8.3964 & 37.9443 & 0.9204 & 0.0314 & 1.5604 \\ \hline
PNN    & 0.0689 & 40.7550 & 0.9624 & 0.0259  & 1.0646 & 29.9418 & 0.9121 & 0.0824 & 3.3206 & 43.1208 & 0.9704 & 0.0172 & 0.8528 \\
PANNET & 0.0688  & 40.8176 & 0.9626 & 0.0257  & 1.0557 & 29.6840  & 0.9072 & 0.0851 & 3.4263 & 43.0659 & 0.9685 & 0.0178 & 0.8577 \\
MSDCNN & 0.2390  & 41.3355 & 0.9664 & 0.0242  & 0.994  & 30.3038 & 0.9184 & 0.0782 & 3.1884 & 45.6874 & 0.9827 & 0.0135 & 0.6389 \\
SRPPNN & 1.7114 & \textcolor{blue}{\ul 41.4538} & 0.9679 & \textcolor{blue}{\ul 0.0233}  & \textcolor{blue}{\ul 0.9899} & \textcolor{blue}{\ul 30.4346} & \textcolor{blue}{\ul 0.9202} & \textcolor{blue}{\ul 0.0770}  & \textcolor{blue}{\ul 3.1553} & \textcolor{blue}{\ul 47.1998} & \textcolor{blue}{\ul 0.9877} & \textcolor{blue}{\ul 0.0106} & \textcolor{blue}{\ul 0.5586} \\
GPPNN  & 0.1198  & 41.1622 & 0.9684 & 0.0244  & 1.0315 & 30.1785 & 0.9175 & 0.0776 & 3.2593 & 44.2145 & 0.9815 & 0.0137 & 0.7361 \\ \hline
Ours   & 0.0873  & \textcolor{red}{\textbf{41.8584}} & \textcolor{blue}{\ul 0.9712} & \textcolor{red}{\textbf{0.0224}}  & \textcolor{red}{\textbf{0.9335}} & \textcolor{red}{\textbf{30.4873}} & \textcolor{red}{\textbf{0.9221}} & \textcolor{red}{\textbf{0.0751}} & \textcolor{red}{\textbf{3.1142}} & \textcolor{red}{\textbf{47.2533}} & \textcolor{red}{\textbf{0.9884}} & \textcolor{red}{\textbf{0.0103}} & \textcolor{red}{\textbf{0.5512}} \\ \hline
\end{tabular}%
}
\vspace{-0.6em}
\end{table*}

\begin{figure*}[ht]
  \centering
  %\fbox{\rule{0pt}{2in} %\rule{0.9\linewidth}{0pt}}
  \includegraphics[width=0.72\linewidth]{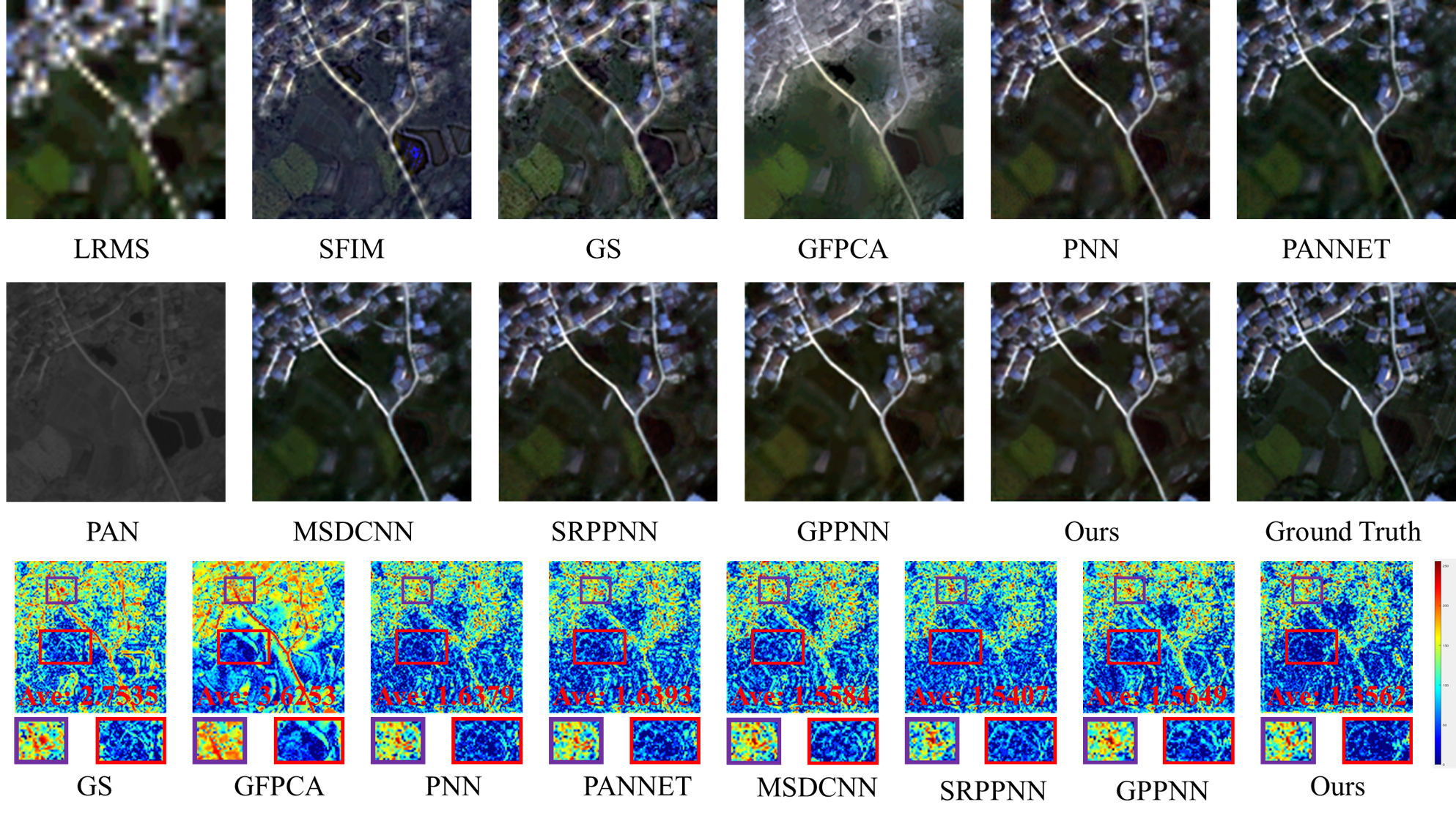}
  \vspace{-0.8em}
  \caption{Visual comparison of all the competing methods on WorldViewII. The last row visualizes the error maps and average errors between the pan-sharpening results and the ground truth.}
  \label{fig:Qualitative_WV2}
  \vspace{-1em}
\end{figure*}

\section{Experiments}
\label{sec:exp}

\subsection{Datasets and evaluation metrics}
\label{sec:dataset}
In this section, we conduct several experiments to verify the effectiveness of our proposed PanFlowNet on three satellite image datasets, i.e., WorldView II, WorldView III, and GaoFen2. For each dataset, we have hundreds of image pairs, and the MS images are cropped into patches with the size of 32 × 32, and the size of corresponding PAN images is 128 × 128. Each patch is normalized into 0 to 1. 

Four assessment metrics are used to evaluate the performance, including peak signal-to-noise ratio (PSNR), Structural similarity (SSIM), Erreur Relative Globale Adimensionnelle de Synthese (ERGAS), and Spectral angle mapper (SAM). The first three metrics measure spatial distortion, and the fourth measures spectral distortion.

\begin{table}[!ht]
  \centering
  \caption{
  % \textbf{The effect of stage.} 
  PSNR values of PanFlowNet with different noises.}
  \label{tab:assess_motivatoin}
  \renewcommand\arraystretch{1.2}
  \resizebox{\columnwidth}{!}{%
  \begin{tabular}{c|cccc}
  \hline
  \rowcolor[HTML]{ECF4FF} 
  {Noise}          & PSNR $\uparrow$            & {SSIM}  $\uparrow$          & {SAM} $\downarrow$            & {ERGAS} $\downarrow$          \\ \hline
  noise 1       & 41.8561          & 0.971218          & 0.0223989          & 0.933770          \\
  noise 2       & 41.8581          & 0.971229         & 0.0223936          & 0.933516          \\
  noise 3       & 41.8579          & 0.971224          & 0.0223922          & 0.933545          \\
  noise 4      & 41.8563           & 0.971216          & 0.0223946          & 0.933642    \\
  noise 5      & 41.8583         & 0.971228          & 0.0223937          & 0.933529    \\
  noise 6      & 41.8552           & 0.971204          & 0.0224002          & 0.933823    \\ \hline
  \end{tabular}%
  }
  \vspace{-0.6em}
  \end{table}

\subsection{Implement details}
We implement our PanFlowNet in the PyTorch framework. As the paired training samples are not available, we construct the paired training datasets using the Wald protocol~\cite{wald1997fusion}. 
To increase training efficiency, we first pre-train our model using an L1 loss for 1000 epochs and then train the whole network using only the loss for 100 epochs. In the training stage, we employ ADAM optimizer with $\beta_1 = 0.9$, $\beta_2 = 0.999$ to update the network parameters for 1000 epochs with a batch size of 4. The learning rate is initialized with $5e-5$ and is decayed by multiplying $0.5$ for every 200 epochs. In the inference stage, we randomly select a Gaussian noise.
All the experiments are conducted on NVIDIA GeForce GTX 3080Ti GPU. 

\subsection{Effectiveness verification}
To verify the effectiveness of the flow-based modelling methodology in our proposed PanFlowNet, we sample different HRMS images from the distribution of HRMS once the PanFlowNet has been trained. Specifically, we use different noise samples to generate the HRMS, and the quantitative experimental results are shown in Table~\ref{tab:assess_motivatoin}. From Table~\ref{tab:assess_motivatoin}, it can be seen that there exist some differences between the restored HRMS. These differences mainly attribute to the different noises. Additionally, we also present the qualitative results in Fig.~\ref{fig:assess_motivation}, from which we can observe that the generated HRMS images from different noises are very similar, but there still exist some differences in their fine details, which indicates that the generated different HRMS images will focus on different detailed parts of the ground-truth\footnote{More results will be presented in the supplementary material.}. 

\subsection{Comparison with the state-of-the-arts}
\label{sec:results}

In this section, to verify the effectiveness of our proposed PanFlowNet, we compare PanFlowNet with ten competitive methods, including five classical methods (i.e., SFIM~\cite{liu2000smoothing}, Brovey~\cite{gillespie1987color}, GS~\cite{laben2000process}, IHS~\cite{haydn1982application}, and GFPCA~\cite{liao2015two}) and five DL-based methods (i.e., PNN~\cite{masi2016}, PANNET~\cite{yang2017pannet}, MSDCNN~\cite{yuan2018multiscale}, SRPPNN~\cite{cai2020super}, and GPPNN~\cite{Xu_2021_CVPR}), which our method is conducted by randomly selected Gaussian noise for inference.

\paragraph{Parameter numbers vs. model performance.}\;
The comparison results between parameter number and model performance are shown in Table~\ref{tab:cmp_sota}, from which it can be seen that our network is able to achieve a good trade-off and perform best with comparably fewer parameters compared to other deep learning-based methods. 

\paragraph{Evaluation on full-resolution scene.}\;
 % In order to demonstrate the real-world application value, we further perform experiments on 200 sets of full-resolution data obtained by Gaofen2. Due to the unavailability of ground-truth MS images in the real-world full-resolution scenes, the commonly-used three non-reference metrics of $D_\lambda$, $D_s$ and $QNR$ are adapted for evaluation.  The quantitative comparison between representative CNN-based methods and our method are shown in Table~\ref{tab:full-image}. The lower $D_\lambda$, $D_s$ and the higher $QNR$ correspond to the better image quality where the best results are remarked by red bold. As can be seen clearly, our methods surpass other competitive Pan-sharpening methods in all the indexes.
 In order to compare the generalization of methods, we further perform experiments on an additional real-world full-resolution dataset of 200 samples obtained by the GaoFen2 satellite for evaluation. Due to the lack of ground-truth MS images in real-world full-resolution scenes, we measure the model's performance using commonly used three non-reference metrics: the spectral distortion index $D_\lambda$, the spatial distortion index $D_s$, and the quality without reference $QNR$. The quantitative comparisons between representative CNN-based methods and our method are shown in Table~\ref{tab:full-image}. The lower $D_\lambda$, $D_s$ and the higher $QNR$ correspond to the better image quality where the best results are remarked by red bold. 
 From Table~\ref{tab:full-image}, our methods surpass other competitive Pan-sharpening methods in all the indexes, which shows its generalization ability.
 % As can be seen clearly, our methods surpass other competitive Pan-sharpening methods in all the indexes.

\begin{table}[t]
% \vspace{-1.2em}
\caption{Non-reference metrics on full-resolution dataset.}
% \vspace{-0.8em}
\centering
\label{tab:full-image}
\resizebox{\linewidth}{!}{%
\begin{tabular}{c|cccccc}
\hline
\rowcolor[HTML]{ECF4FF} 
      & PAN    & PANNET & MSDCNN & SRPPNN  & GPPNN  & Ours  \\ \hline
$D_{\lambda}$ $\downarrow$ & 0.0746 & 0.0737 & \textcolor{blue}{\ul 0.734} & 0.0767 & 0.0782 & \textcolor{red}{\textbf{0.0665}} \\
$D_s$ $\downarrow$         & 0.1164 & 0.1224 & \textcolor{blue}{\ul 0.1151} & 0.1162 & 0.1253 & \textcolor{red}{\textbf{0.1113}} \\
QNR $\uparrow$          & 0.8191 & 0.8143 & \textcolor{blue}{\ul 0.8215} & 0.8173 & 0.8073 & \textcolor{red}{\textbf{0.8257}} \\ \hline
\end{tabular}%
}
\vspace{-1.6em}
\end{table}

\noindent \textbf{Quantitative results.}\;
The comparison results of 10 benchmark methods over three satellite datasets are reported in Table~\ref{tab:cmp_sota}, where the best and the second best values are highlighted in red bold and blue underline, respectively. As can be seen clearly, our proposed method achieves the best overall results over other competing methods over all the satellite datasets. This confirms, to a certain extent, the effectiveness and flexibility of our method.

\noindent \textbf{Qualitative results.}\;
We also show the visual results of one image from the WorldView II dataset in Fig.~\ref{fig:Qualitative_WV2} to evaluate the effectiveness of our method. From the first two rows of Fig.~\ref{fig:Qualitative_WV2}, we can see that the visual result of our PanFlowNet is obviously better than other competing methods. To make the visual advantage clear, the images in the last row are the error maps and the average errors between the output pan-sharpened results and the ground truth. Compared with other competing methods, our PanFlowNet has the minimum spatial and spectral distortions. As for the error maps, it is noted that our proposed method has the smallest average error compared to other comparison methods while being the closest to ground truth. The state-of-the-art performance of our method demonstrates the effectiveness of the proposed PanFlowNet\footnote{ More qualitative results are presented in the supplementary material.}.

\begin{table}[t]
\centering
\caption{
% \textbf{The effect of stage.} 
PSNR values of PanFlowNet with different number of stages on WorldViewII. The best and the second best values are highlighted in \textcolor{red}{\textbf{bold}} and \textcolor{blue}{\ul underline}, respectively. }
\label{tab:stage}
\renewcommand\arraystretch{1.1}
\resizebox{\columnwidth}{!}{%
\begin{tabular}{c|cccc}
\hline
\rowcolor[HTML]{ECF4FF} 
{Stages (K)}          & PSNR $\uparrow$            & {SSIM}  $\uparrow$          & {SAM} $\downarrow$            & {ERGAS} $\downarrow$          \\ \hline
1       & 38.2469          & 0.9471          & 0.0344          & 1.4294          \\
2      & 40.7152           & 0.9639          & 0.0255          & 0.9935    \\
3      & \textcolor{blue}{\ul 41.2664}          & \textcolor{blue}{\ul 0.9674}          & \textcolor{blue}{\ul 0.0236}          & \textcolor{blue}{\ul 0.9935}          \\
4      & \textcolor{red}{\textbf{41.8584}} & \textcolor{red}{\textbf{0.9712}} & \textcolor{red}{\textbf{0.0224}} & \textcolor{red}{\textbf{ 0.9335}} \\ \hline
\end{tabular}%
}
\vspace{-1.2em}
\end{table}

\begin{table*}[h!t]
\centering
\caption{The results of different configurations on WorldViewII. The best and the second best values are highlighted in \textcolor{red}{\textbf{bold}} and \textcolor{blue}{\ul underline}, respectively. (PS: Parameters Sharing)}
\vspace{-0.4em}
\label{tab:ablation}
\renewcommand\arraystretch{1.1}
\resizebox{0.68\textwidth}{!}{%
\begin{tabular}{c|ccc|cccc}
\hline
\rowcolor[HTML]{ECF4FF} 
Configuration &  L & P & PS & PSNR $\uparrow$    & SSIM $\uparrow$  & SAM $\downarrow$   & ERGAS $\downarrow$ \\ \hline
I             & \XSolid    & \XSolid  & \Checkmark  & 31.3136 & 0.9033 & 0.0840 & 3.2813 \\
II            & \Checkmark  & \XSolid    & \Checkmark        & 36.1760 & 0.9058 & 0.0315 & 1.6287 \\
III            &\XSolid   & \Checkmark     & \Checkmark        & 40.8503 & 0.9647 & 0.0253 & {1.0539} \\  \hline
IV            & \Checkmark  & \Checkmark  & \XSolid        & \textcolor{red}{\textbf{42.0865}} & \textcolor{red}{\textbf{0.9719}} & \textcolor{red}{\textbf{0.0215}} & \textcolor{red}{\textbf{0.9062}} \\ \hline
PanFlowNet(Ours)          & \Checkmark   & \Checkmark    & \Checkmark       &  \textcolor{blue}{\ul 41.8584} & \textcolor{blue}{\ul 0.9712} & \textcolor{blue}{\ul 0.0224} & \textcolor{blue}{\ul 0.9335} \\ \hline
\end{tabular}%
}
% \vspace{-1.1em}
\end{table*}

\begin{figure*}[ht]
    \centering
    \includegraphics[width=0.68\linewidth]{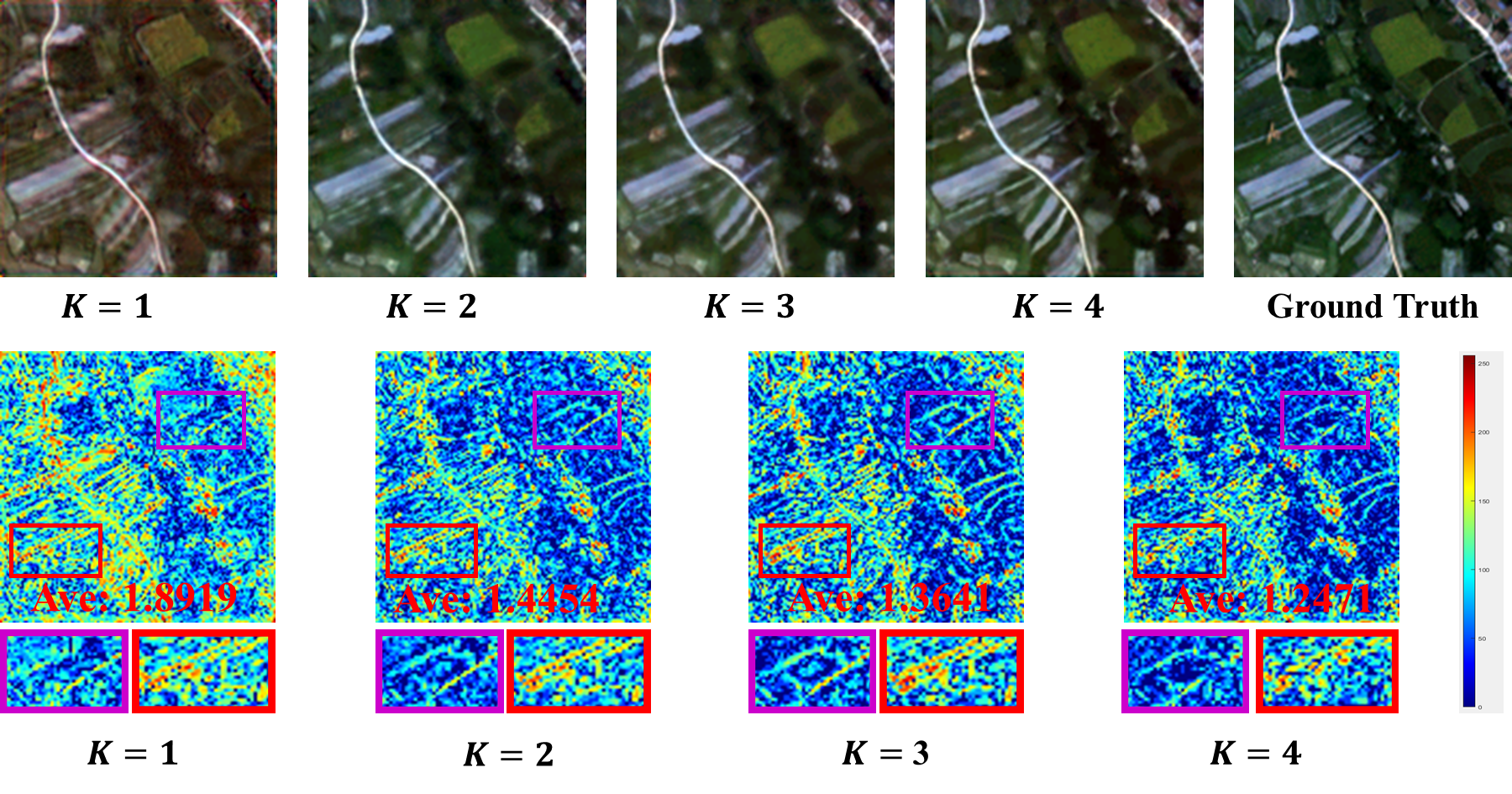}
    \vspace{-0.8em}
    \caption{Intermediate visual results of different numbers of CACB in our PanFlowNet on WorldViewII. The last row visualizes the error maps and average errors between the pan-sharpening results and the ground truth.}
    \label{fig:Qualitative_inter}
    \vspace{-1.2em}
\end{figure*}

\subsection{Ablation study}
\label{sec:ablation}

We conduct ablation studies to further validate the effectiveness of our model under different configurations, including different numbers of CACB in the model, different condition settings, and different parameter sharing settings.

\noindent \textbf{The number of CACB.}\;
To explore the impact of the number of CACB (i.e., K) in our PanFlowNet, we conduct the experiment with a varying number of parameters K. Table \ref{tab:stage} shows the results of different K from 1 to 4. It can be seen that the PSNR performance increases as the number of stages increases. For easy observation, we also visualize obtained HRMS images for different numbers of CACB in Fig.~\ref{fig:Qualitative_inter}, from which we can see that the visual results and the error maps are the best when the number of CACB is $4$. Thus we choose $K=4$ in all the experiments. 
% to balance the performance and computational complexity.

\noindent \textbf{Influence of different condition.}\;
In the pan-sharpening task, two conditions exist for our method
($\mathbf{L}$ and $\mathbf{P}$), and here, we employ an ablation study to explore the effectiveness of these two conditions. As shown in Tab.~\ref{tab:ablation}, we can see that the best performance improvement is achieved when both conditions are available, while the absence of any condition leads to the worst performance due to the fact that it becomes a purely generative task.

\noindent \textbf{Parameter sharing.}\;
We evaluate the scenario where the parameters are not shared when $K=16$. In other words, the parameters of different CABAs in PanFlowNet are no longer shared. From Table~\ref{tab:ablation}, we can see that the performance of the model will be improved to some extent without parameter sharing, but it is not a good choice for reducing the model complexity. In our experiment, we still adopt the parameter sharing technique.
% \blindtext

% \subsubsection{Feature dimension}

% \blindtext

\subsection{Limitation}
\label{sec:limitation}

Our work has several limitations. The generated HRMS images are diverse, and different HRMS images with different properties are sampled. For this diversity, our method has weak controllability for such different properties of HRMS images and cannot readily generate the HRMS images that match our desired properties, such as generating HRMS images with higher SSIM. In future work, we will try to add controllable elements to control the generated HRMS images so that they can satisfy our demand.
%In addition, we choose the standard Gaussian distribution for the hidden variable distribution, but if we cannot transform to this distribution in the transformation, it will produce a gap, so we need to choose simple known distributions appropriately for different image distributions.

\section{Conclusion}
\label{sec:conclusion}

In this paper, we proposed a  novel neural network architecture used for pan-sharpening called PanFlowNet. Specifically, we introduce a flow-based deep network for pan-sharpening, and this network is capable of accurately learning the distribution of realistic HRMS images condition on the LRMS and PAN images. To the best of our knowledge, this is the first attempt to employ generative methods to learn the distribution of HRMS samples for the pan-sharpening task. The trained network can generate diverse HRMSs by inputting different noises. Extensive experimental results demonstrate the effectiveness and superiority of the proposed network.

{\small
\bibliographystyle{ieee_fullname}
\bibliography{main}
}

\end{document}